\documentclass[10pt,twocolumn,letterpaper]{article}

\usepackage[pagenumbers]{cvpr} 

\usepackage{graphicx}
\usepackage{amsmath}
\usepackage{amssymb}
\usepackage{booktabs}
\usepackage{multirow}
\newcommand\blfootnote[1]{%
  \begingroup
  \renewcommand\thefootnote{}\footnote{#1}%
  \addtocounter{footnote}{-1}%
  \endgroup
}

\usepackage[pagebackref,breaklinks,colorlinks]{hyperref}

\usepackage[capitalize]{cleveref}
\crefname{section}{Sec.}{Secs.}
\Crefname{section}{Section}{Sections}
\Crefname{table}{Table}{Tables}
\crefname{table}{Tab.}{Tabs.}

\def\cvprPaperID{*****} 
\def\confName{CVPR}
\def\confYear{2022}

\begin{document}

\title{MV-FCOS3D++: Multi-View Camera-Only 4D Object Detection \\with Pretrained Monocular Backbones}

\author{Tai Wang$^{1*\dagger}$\quad Qing Lian$^{2*}$\quad Chenming Zhu$^3$\quad Xinge Zhu$^1$\quad Wenwei Zhang$^4$\vspace{1ex} \\
{\small$^1$The Chinese University of Hong Kong\quad $^2$Hong Kong University of Science and Technology}\\
{\small$^3$The Chinese University of Hong Kong, Shenzhen\quad $^4$Nanyang Technological University}\\
{\tt\small\{taiwang.me, zhuxinge123\}@gmail.com, qlianab@connect.ust.hk}\\
{\tt\small chenmingzhu@link.cuhk.edu.cn, wenwei001@ntu.edu.sg}
}
\maketitle
\blfootnote{$^*$ Equal contribution.\ \  $\dagger$ Corresponding author.}

\begin{abstract}
In this technical report, we present our solution, dubbed MV-FCOS3D++, for the Camera-Only 3D Detection track in Waymo Open Dataset Challenge 2022.
For multi-view camera-only 3D detection, methods based on bird-eye-view or 3D geometric representations can leverage the stereo cues from overlapped regions between adjacent views and directly perform 3D detection without hand-crafted post-processing.
However, it lacks direct semantic supervision for 2D backbones, which can be complemented by pretraining simple monocular-based detectors.
Our solution is a multi-view framework for 4D detection following this paradigm.
It is built upon a simple monocular detector FCOS3D++, pretrained only with object annotations of Waymo, and converts multi-view features to a 3D grid space to detect 3D objects thereon.
A dual-path neck for single-frame understanding and temporal stereo matching is devised to incorporate multi-frame information.
Our method finally achieves 49.75\% mAPL with a single model and wins 2nd place in the WOD challenge, without any LiDAR-based depth supervision during training.
The code will be released at \footnotesize\url{https://github.com/Tai-Wang/Depth-from-Motion}.
\end{abstract}
\vspace{-2ex}
\section{Introduction}
The Waymo Open Dataset Challenge at CVPR 2022 is one of the largest and most challenging competitions for autonomous driving. This year, it sets up a new track for camera-only 3D detection, which requires the algorithm to localize and classify 3D objects given only images from multiple cameras.
In contrast to previous benchmarks~\cite{KITTI,nuScenes} and the original Waymo 3D detection track, this challenge provides user-friendly camera synced labels for training and proposes a custom metric, LET-3D-APL, to evaluate camera-only 3D detectors. Both make this challenge a promising benchmark that encourages new insights and methods in this stream.

In this challenge, motivated by ImVoxelNet~\cite{imvoxelnet}, we explore a general solution, MV-FCOS3D++, built upon an explicit 3D voxel representation for performing multi-view camera-only 3D detection. This pre-defined voxel grid provides a unified, regular structure bridging the monocular features from different views and serves as the volume space for temporal stereo matching.
It enables the framework to conduct multi-view 3D detection in a simple and unified manner but lacks the direct image-view semantic supervision for 2D feature extraction.
To address this issue, we pretrain the 2D backbone based on a simple monocular 3D detector, FCOS3D++\footnote{The manuscript of FCOS3D++ will be released soon.}~\cite{FCOS3D,pgd}, with only object annotations on Waymo. It enhances the backbone's capability of understanding semantics and geometry in monocular images and improves the 3D detection performance significantly.
Furthermore, we devise a dual-path scheme to incorporate multi-frame information in our main framework.
It disentangles single-frame understanding from temporal stereo matching and naturally compensates the latter for cases when temporal matching breaks down, such as static scenes and moving objects.

Our method finally achieves 49.75\% mAPL with a single model and wins 2nd place in the WOD challenge, without any LiDAR-based depth supervision during training.

\begin{figure*}
    \centering
    \includegraphics[width=1.0\textwidth]{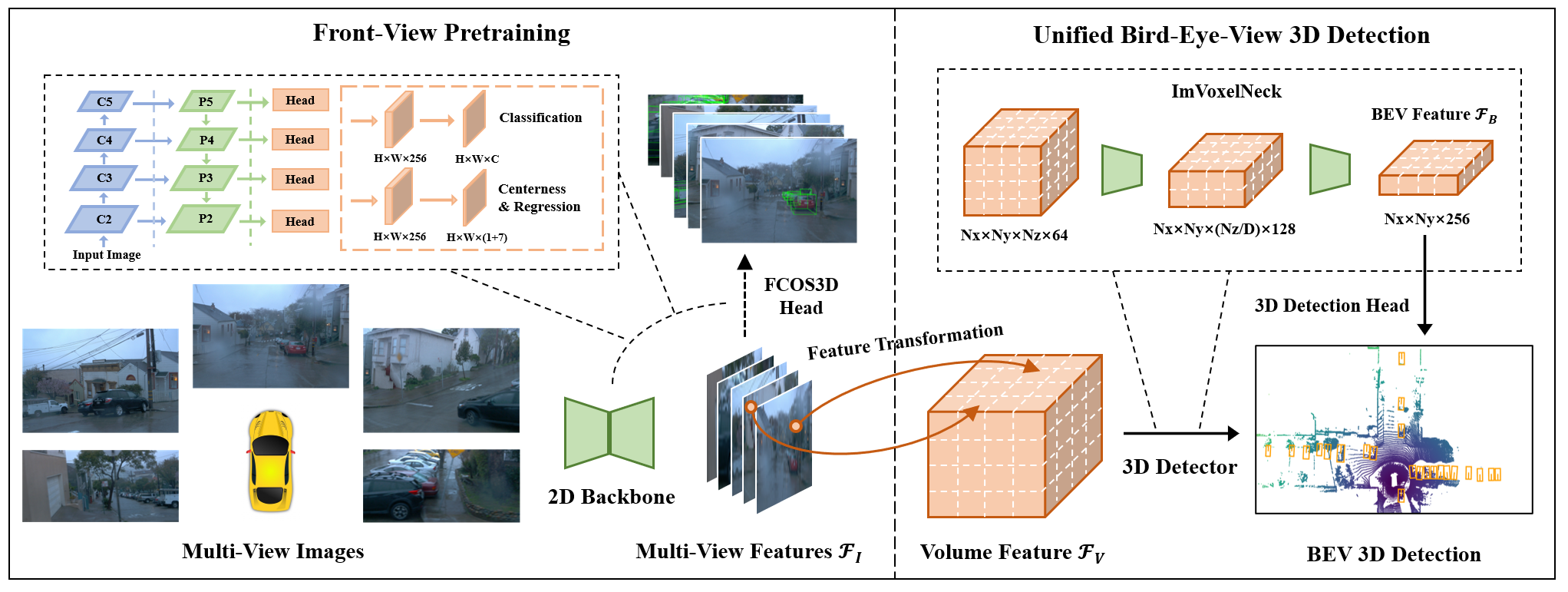}
    \vspace{-4ex}
    \caption{An overview of our framework.}
    \vspace{-3ex}
    \label{fig:overview}
\end{figure*}
\section{Methodology}
This section will introduce the details of our winning solution, MV-FCOS3D++ (Fig.~\ref{fig:overview}). We first give the overview of our framework from the perspective of 2D feature extraction, feature transformation from perspective view (FOV) to bird-eye-view (BEV), and the subsequent simple 3D detector. Then we introduce the pre-training and temporal modeling techniques that enhance our MV-FCOS3D++. 
\subsection{MV-FCOS3D++}\label{sec:overview} 
\noindent\textbf{2D Feature Extraction.}\quad
Given a 2D image $\mathcal{I} \in \mathcal{R}^{W\times H \times 3}$, the feature extraction module aims at extracting its high-level information to understand the semantics and geometry structure therein. Following \cite{FCOS3D, imvoxelnet}, we use a shared 2D backbone ResNet-101 with DCN~\cite{DeformConv} to extract the features and aggregate the multi-scale features by Feature Pyramid Network (FPN) to get $\mathcal{F}_{I} \in \mathcal{R}^{\frac{W}{4} \times \frac{H}{4} \times c}$ (P2 feature level in Fig.~\ref{fig:overview}) for each view of image.

\noindent\textbf{Feature Transformation.}\quad
Given the FOV features $\mathcal{F}_{I}$ of multiple views, the feature transformation module uses the camera intrinsic to lift it to 3D space and obtains a unified volume feature $\mathcal{F}_{V} \in R^{N_x \times N_y \times N_z \times c}$, where $N_x$, $N_y$ and $N_z$ denote the grid size in the $x, y, z$ axis. Specifically, we first pre-define the 3D voxel space and grid sample each point to construct the 3D volume feature. 
For each point $(x, y, z)$, the corresponding 3D features $\mathcal{F}_V (x, y, z)$ is obtained via:
\begin{align}
    \mathcal{F}_V\left(x, y, z\right) = \mathcal{F}_{I}\left( \pi(x, y, z) \right),
\end{align}
where $\pi$ denotes the 3D to 2D coordinate projection (without considering rolling shutter for simplicity). Since the driving cars in the Waymo dataset are equipped with surround cameras, one 3D point may correspond to multiple 2D points in the images. To handle this situation, we adopt the mean average pooling to aggregate the features from multiple 2D points. 

\noindent\textbf{Voxel-Based 3D Detector.}\quad
After obtaining the 3D voxel feature, we utilize several residual blocks composed of 3D convolutional neural networks following ImVoxelNet~\cite{imvoxelnet} to aggregate the 3D spatial information and compress it along z-axis to get the BEV features $\mathcal{F}_{B} \in \mathcal{R}^{N_x \times N_y \times 4c}$. 

With the BEV features $\mathcal{F}_B$, we follow BEV-based detectors~\cite{imvoxelnet,CaDDN,PointPillars} and conduct 3D detection in the 2D space. In this competition, we mainly study two kinds of detection heads: anchor-based 3D head~\cite{PointPillars} and anchor-free, center-based 3D head~\cite{CenterPoint}.

The \textbf{anchor-based 3D head}, a single shot multi-box detector (SSD)-like~\cite{SSD} architecture is also widely used in LiDAR-based 3D object detection~\cite{SECOND, PointPillars}. It consists of three components:  anchor classification, bounding box regression, and direction classification. Anchor classification identifies the positive anchors and estimates the corresponding semantic classes (\textit{i.e.} car, pedestrian, and cyclist). Following~\cite{SECOND, imvoxelnet}, the positive anchors are determined by the IoU between anchors with ground truth in the BEV space. 
The positive and negative thresholds are set to 0.6 and 0.45 for car while 0.5 and 0.35 for pedestrian and cyclist.
During training, the loss in anchor-based head is defined as follows:
\begin{align}
    \mathcal{L} = \mathcal{L}_{cls} + \lambda_{reg}\mathcal{L}_{reg} + \lambda_{dir}\mathcal{L}_{dir},
\end{align}
where $\mathcal{L}_{cls}$ denotes the focal loss for anchor classification, $\mathcal{L}_{reg}$ denotes the smooth L1 loss for the bounding box regression, $\mathcal{L}_{dir}$ denotes the cross entropy loss for direction classification, $\lambda_{reg}$ and $\lambda_{dir}$ are 2 and 0.2, respectively.
During inference, we filter the redundant prediction through Non-Maximum Suppression (NMS), where the redundant criterion is based on the IoU in the BEV space. The number of predictions before and after NMS is set to 4096 and 500 to increase the recall performance.

The \textbf{center-based 3D head}~\cite{CenterPoint} is an anchor-free detection head. It first localizes the object center based on the keypoint network and then regresses the bounding box attributes.
The regression part consists of the object location, dimension and the cosine and sine value of bounding box yaw angle. The overall loss is defined as follows:
\begin{align}
    \mathcal{L} = \mathcal{L}_{key} + \lambda_{reg}\mathcal{L}_{reg},
\end{align}
where $ \mathcal{L}_{key}$ denotes the gaussian-based focal loss for keypoint localization, $\mathcal{L}_{reg}$ represents the L1 loss for the regression part and $\lambda_{reg}$ is 0.25.
During inference, we utilize the pooling-based peak keypoint extraction to obtain the object centers and construct 3D bounding boxes.

\subsection{Pretraining with Perspective-View Supervision}
As observed in~\cite{DETR3D, CaDDN}, BEV-based 3D detectors can benefit from backbone pretraining with monocular-based paradigms due to the lack of perspective-view supervision.
To this end, we first pretrain our 2D feature extraction components by a simple monocular-based 3D detector, FCOS3D++~\cite{FCOS3D,pgd}, with \emph{only object annotations} and then finetune them with a smaller learning rate (0.1$\times$) when training the detector in the BEV space subsequently.
The implementation of FCOS3D++ follows its open-source version~\cite{mmdet3d2020} while adjusting the depth and 3D size priors according to the statistics on Waymo. Besides, we only use P3-P5 (Fig.~\ref{fig:overview}) with regression ranges set to (0, 128, 256, $\infty$) to produce multi-level predictions more efficiently.

So far, we have set up a baseline that can conduct 3D detection from single-frame multi-view images. Although it has incorporated a few stereo cues lying in the overlapped regions between adjacent views, they are still limited for estimating object depths accurately. Next, we will show how we exploit the stereo cues provided by consecutive frames.

\subsection{Dual-Path Temporal Modeling}~\label{sec:temporal}
Similar to typical multi-view or binocular settings, two images nearby in temporal also have stereo correspondence for a static environment.
Compared to monocular-based understanding, the underlying philosophies of stereo depth estimation are different: it relies on matching instead of data-driven monocular priors. Therefore, we use \emph{concatenation} instead of simple average pooling to construct multi-frame volumetric features.
In addition, although stereo estimation can leverage the strong cues provided by absolute ego motions, there are multiple cases that stereo estimation approaches can not handle, such as static scenes and moving objects. So we further devise a dual-path scheme to keep the monocular understanding branch and allow it to adaptively compensate the stereo estimation.

\begin{figure}
    \centering
    \includegraphics[width=1.0\linewidth]{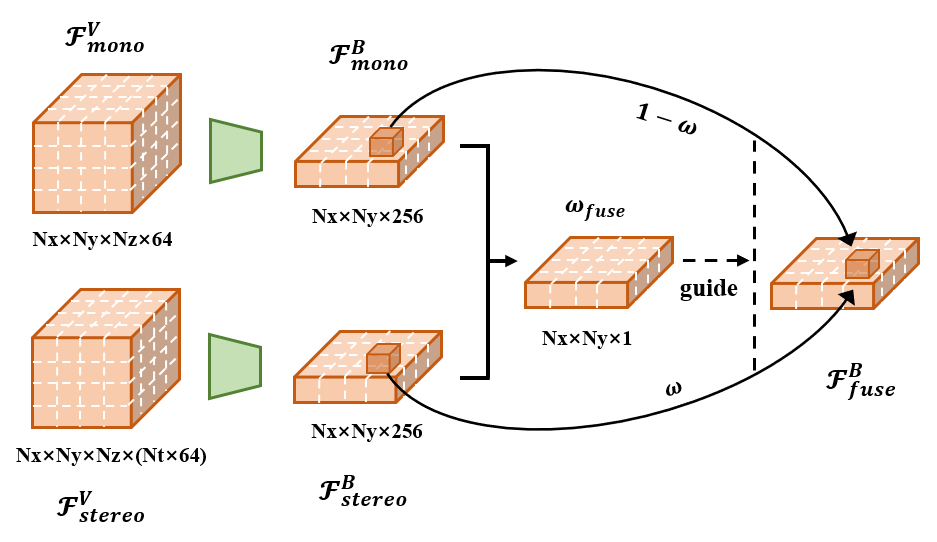}
    \vspace{-3ex}
    \caption{Our dual-path design for temporal modeling.}
    \vspace{-3ex}
    \label{fig:dual-path}
\end{figure}

Formally, as shown in Fig.~\ref{fig:dual-path}, given volumetric features extracted from consecutive frames and transformed into the ego coordinate system in the current frame, we concatenate them along the feature channel to obtain $\mathcal{F}_{stereo}^V$. Then we use two ImVoxelNeck to aggregate monocular ($\mathcal{F}_{mono}^V$) and stereo features ($\mathcal{F}_{stereo}^V$) separately. The network for the stereo path shares the same architecture with the other, except that the input channel is multiplied by the number of frames $N_t$. Then we have two BEV features with the same shape. To fuse these features, we concatenate them and feed them into a simple 2D convolutional layer composed of 1$\times$1 kernel, and aggregate it along the feature channel to get a point-wise weight feature map. Then the sigmoid response of this feature serves as the weight $\omega_{fuse}$ for guiding the fusion of $\mathcal{F}_{mono}^B$ and $\mathcal{F}_{stereo}^B$. Denoting the convolutional network as $\phi$, this procedure is represented as follows:
\begin{equation}
    \omega_{fuse}=\sigma(\phi(\mathcal{F}_{mono}^B,\mathcal{F}_{stereo}^B)),
\end{equation}
\begin{equation}
    \mathcal{F}_{fuse}^B=\omega_{fuse}\circ \mathcal{F}_{stereo}^B+(1-\omega_{fuse})\circ \mathcal{F}_{mono}^B.
\end{equation}
Here $\sigma$ denotes the sigmoid function, and $\circ$ refers to element-wise multiplication. The derived stereo feature $\mathcal{F}_{fuse}^B$ is finally input to the subsequent 3D detection head.

In practice, to avoid much memory overhead, we cut off the gradient back-propagation of the 2D backbone for previous frames and only sample one from the previous ten frames to construct stereo pairs. During inference, we use the most previous one (at most the previous 10th frame) to guarantee relatively large view changes. When there is no previous frame, we copy the current frame as the previous frame and the weight can empirically switch to relying on the monocular path. This training and inference setting empirically perform best, surpassing other choices (5 or 20 frames) by 0.5\% mAPL.
\begin{table*}[htbp]
  \small
  \centering
  \caption{Main results. "full res." and "w/ pt." represent that the input has full resolution and the backbone is initialized with weights pretrained from FCOS3D++, respectively. Note that "+" denotes the orthogonal modifications based on MV-FCOS3D++ (w/ pt.). The cyclist performance of FCOS3D++ is much lower possibly because the strong uncertainty filters out too many positive predictions.}
  \vspace{-1ex}
    \begin{tabular}{l|r|r|cccccc|cc} \hline
    \multicolumn{1}{c|}{\multirow{2}[0]{*}{Method }} & \multicolumn{1}{c|}{\multirow{2}[0]{*}{Split}} & \multicolumn{1}{c|}{\multirow{2}[0]{*}{Data}} & \multicolumn{2}{c}{Car} & \multicolumn{2}{c}{Pedestrian} & \multicolumn{2}{c}{Cyclist} & \multicolumn{2}{|c}{Average} \\ 
          & & & APL   & AP    & APL   & AP    & APL   & AP    & APL   & AP \\\hline
    FCOS3D++ (full res.) & \multicolumn{1}{c|}{\multirow{3}[0]{*}{Val}} & \multicolumn{1}{c|}{\multirow{3}[0]{*}{1/5}} & 40.48 & 57.36	& 21.48 & 32.92 & 2.02 & 3.12	& 21.33	& 31.13 \\
    DETR3D (vanilla, full res.) & & & 34.61 & 48.09 & 25.61 & 38.56 & 18.01 & 30.34 & 26.08 & 39.00 \\
    DETR3D (w/ pt., full res.) & & & 39.9  & 53.05 & 29.34 & 43.94 & 20.01 & 31.87 & 29.75 & 42.95 \\
    \hline
    MV-FCOS3D++ (Baseline) & \multicolumn{1}{c|}{\multirow{5}[0]{*}{Val}} &  \multicolumn{1}{c|}{\multirow{5}[0]{*}{1/5}} & 47.27 & 62.81 & 14.22	& 21.16 & 12.44	& 18.96 & 24.65	& 34.31 \\
    MV-FCOS3D++ (w/ pt.) & & &  50.66 & 66.34 & 20.93 & 30.47 & 18.12 & 26.82 & 29.90 & 41.21 \\
    + adjust assign params & & &  53.90	 & 69.55 & 28.59 & 40.70 & 24.18 & 36.75 & 35.56 & 49.00\\
    + adjust assign \& full res. \& temporal & & & 58.30 & 73.63 & 30.98 & 43.69 & 25.64 & 36.90 & 38.31 & 51.41 \\
    + adjust assign \& infer. params & & & 57.33 & 74.16 & 36.11 & 50.87 & 25.04 & 37.83 & 39.50 & 54.29  \\
    + use CenterHead & & & 53.28 & 69.74 & 36.87 & 52.39 & 26.23 & 39.55 & 38.79 & 53.89 \\
    \hline\hline
    3DMVT & \multicolumn{1}{c|}{\multirow{2}[0]{*}{Test}} & \multicolumn{1}{c|}{\multirow{2}[0]{*}{-}} &  60.89 & 78.01 & 44.83 & 63.85 & 40.56 & 54.66 & 48.76 & 65.50 \\
    BEVFormer~\cite{bevformer} & & & 68.78 & 82.91 & 53.20 & 71.05 & 46.50 & 58.10 & 56.16 & 70.69 \\ \hline
    MV-FCOS3D++ (single model) & \multicolumn{1}{c|}{\multirow{2}[0]{*}{Test}} & \multicolumn{1}{c|}{\multirow{2}[0]{*}{full}} & 63.90 & 79.18 & 44.72 & 62.67 & 40.63 & 51.92 & 49.75 & 64.59\\
    MV-FCOS3D++ (ensemble) & & & 66.8  & 82.01 & 45.15 & 63.14 & 41.87 & 53.66 & 51.27 & 66.27 \\ \hline
    \end{tabular}%
  \vspace{-1ex}
  \label{tab: main-results}%
\end{table*}%

\begin{table*}[htbp]
  \small
  \centering
  \caption{Quantitative analysis for the performance of different camera views}
  \vspace{-1ex}
    \begin{tabular}{l|cccccccccc|cc} \hline
    \multicolumn{1}{c|}{\multirow{2}[0]{*}{Method }} & \multicolumn{2}{c}{Front} & \multicolumn{2}{c}{Front-Left} & \multicolumn{2}{c}{Front-Right} & \multicolumn{2}{c}{Side-Left} & \multicolumn{2}{c}{Side-Right} & \multicolumn{2}{|c}{Average} \\ 
          & mAPL & mAP & mAPL & mAP & mAPL & mAP & mAPL & mAP & mAPL & mAP & mAPL & mAP \\\hline
    3DMVT &  48.74 & 64.65 & 50.19 & 66.50 & 50.74 & 69.02 & 42.84 & 58.28 & 41.19 & 58.71 & 48.76 & 65.50 \\
    BEVFormer~\cite{bevformer} & 58.21 & 71.29 & 56.51 & 70.41 & 57.13 & 73.92 & 47.47 & 62.80 & 48.39 & 63.53 & 56.16 & 70.69 \\ \hline
    Ours (single) & 48.38 & 61.30 & 51.76 & 66.91 & 52.41 & 69.51 & 44.53 & 59.54 & 47.13 & 62.31 & 49.75 & 64.59\\
    Ours (ensemble) & 49.39 & 62.73 & 53.10 & 68.34 & 54.49 & 71.11 & 47.23 & 62.75 & 48.64 & 63.60 & 51.27 & 66.27 \\ \hline
    \end{tabular}%
  \vspace{-2ex}
  \label{tab: diff-views}%
\end{table*}%

\section{Experiments}
\subsection{Dataset and Evaluation metrics}
\textbf{Waymo Open Dataset}~\cite{Sun_2020_CVPR} consists of 798/202/80 sequences for training/validation/testing in the camera-only 3D detection track, where each sequence contains 171-200 frames. Regarding to the image data, each frame contains five surround-view images with resolution of $1920\times 1280$ or $1920 \times 886$ pixels. 

In this competition, the longitudinal error tolerant IoU (LET-IoU) based average precision (LET-3D-AP) and average precision weighted by localization affinity (LET-3D-APL) metrics are adopted. Specifically, LET-IoU measures the 3D IoU between the ground truth and the prediction that is corrected with the longitudinal localization error. Based on LET-IoU, LET-3D-AP measures the mean average precision of predictions through their bipartite matching with ground truths. On this basis, LET-3D-APL further considers the localization performance by multiplying the precision in LET-3D-AP with the longitudinal affinity~\cite{hung2022let3dap}.

\subsection{Implementation details}
\noindent\textbf{Training parameters.}\quad
Our training is split into two stages. The first stage is trained with a monocular-based model (FCOS3D++), and the second stage is trained with our BEV-based model. 
In the first stage, we initialize the FOV backbone with the ImageNet-pretrained weights and optimize it with the SGD optimizer. We train the model for 24 epochs with the learning rate of 0.001 and first warm-up in the 500 iterations, then decay after 16, 22 epochs with the ratio of 0.1. 
In the second stage, we further train the BEV model for 24 epochs based on the weights trained in the first stage. We adopt the AdamW optimizer with the step decay learning rate policy and initialize the learning rate as 4e-3 and weight decay as 1e-4. 
The 3D voxel size is (0.5m, 0.5m, 0.5m) and the detection range is [-35m, 75m], [-75m, 75m], [-2m, 4m] for the X, Y and Z axis, respectively.

By default, we use a lower image resolution $1248\times 832$ and only 1/5 training data for ablation studies while using the raw image with $1920\times 1280$ resolution and full dataset in our final models.
Due to the large scale of the Waymo dataset, we only train 12 epochs for the second stage when using the full training split.
The batch size is set as 96 and 32 in the first and the second stage, respectively.

\noindent\textbf{Data augmentation.}\quad
We adopt the FOV-based data augmentation techniques in our BEV-based main framework, including random scaling, random cropping, and random flipping. The implementation of random scaling and flipping follows the standard pipeline in FCOS3D++. The rescaling range is set to $[0.95, 1.05]$. The cropping size is set to $1080\times 720$ and $1536\times 1024$ in the ablation studies and our final experiments, respectively. The random cropping saves a lot of memory, especially in our final models, while can keeping the performance is not affected.



\subsection{Results}
We first show the results of several baselines and key results of our method in Tab.~\ref{tab: main-results}.
Our MV-FCOS3D++ baseline uses ImageNet pretrained backbone and is further enhanced significantly with pretraining FCOS3D++ on Waymo, especially for small objects.
It can be seen that our method shows great superiority compared to the other two baselines, FCOS3D++ and DETR3D.
Then we adjust the assignment settings, \emph{i.e.}, tune the matching IoU thresholds to 0.6/0.45 for car and 0.5/0.35 for pedestrian and cyclist, such that MV-FCOS3D++ can achieve much better performance than FCOS3D++ and DETR3D in each category, even with lower image resolution.
Full resolution and temporal modeling (Sec.~\ref{sec:temporal}) can further enhance our baseline significantly (each contributes about half from 35.56\% to 38.31\% mAPL).
Apart from the hyper-parameters for target assignment during training, increasing the number of predictions before and after NMS to 4096 and 500 is also an important setting during inference. It can improve the recall performance remarkably.
Finally, the center-based 3D detection head shows better performance on small objects, pedestrians and cyclists.

On the test set, our final \emph{single model} result in Tab.~\ref{tab: main-results} is achieved with our center-based model, because we observe that the center-based paradigm can perform much better on small objects, thus showing better performance on average.
The ensemble model uses NMS to merge the predictions of an anchor-based model and a center-based model for three classes and another anchor-based model for only car to get the final result.\footnote{The entry namely \emph{MV-FCOS3D++} and \emph{FCOS3D-MVDet3D} only differs in terms the inference setting of center-based model. The former uses circle-nms~\cite{CenterPoint} to better filter overlapped predictions.}

We also compare our method with others on the performance of different categories (Tab.~\ref{tab: main-results}) and different camera views (Tab.~\ref{tab: diff-views}). We observe that our method performs better on car and side views, which are comparable to or surpass BEVFormer. However, the detection performance of small objects and the front-view drags down the overall result. Future work can be focused on improving these two aspects.

{\small
\bibliographystyle{ieee_fullname}
\bibliography{egbib}

\begin{thebibliography}{10}\itemsep=-1pt

\bibitem{nuScenes}
Holger Caesar, Varun Bankiti, Alex~H. Lang, Sourabh Vora, Venice~Erin Liong,
  Qiang Xu, Anush Krishnan, Yu Pan, Giancarlo Baldan, and Oscar Beijbom.
\newblock nuscenes: A multimodal dataset for autonomous driving.
\newblock {\em CoRR}, abs/1903.11027, 2019.

\bibitem{mmdet3d2020}
MMDetection3D Contributors.
\newblock {MMDetection3D: OpenMMLab} next-generation platform for general {3D}
  object detection.
\newblock \url{https://github.com/open-mmlab/mmdetection3d}, 2020.

\bibitem{DeformConv}
Jifeng Dai, Haozhi Qi, Yuwen Xiong, Yi Li, Guodong Zhang, Han Hu, and Yichen
  Wei.
\newblock Deformable convolutional networks.
\newblock In {\em IEEE International Conference on Computer Vision}, 2017.

\bibitem{KITTI}
Andreas Geiger, Philip Lenz, and Raquel Urtasun.
\newblock Are we ready for autonomous driving? the kitti vision benchmark
  suite.
\newblock In {\em IEEE Conference on Computer Vision and Pattern Recognition},
  2012.

\bibitem{hung2022let3dap}
Wei-Chih Hung, Henrik Kretzschmar, Vincent Casser, Jyh-Jing Hwang, and Dragomir
  Anguelov.
\newblock Let-3d-ap: Longitudinal error tolerant 3d average precision for
  camera-only 3d detection.
\newblock {\em arXiv preprint arXiv:2206.07705}, 2022.

\bibitem{PointPillars}
Alex~H. Lang, Sourabh Vora, Holger Caesar, Lubing Zhou, Jiong Yang, and Oscar
  Beijbom.
\newblock Pointpillars: Fast encoders for object detection from point clouds.
\newblock In {\em IEEE Conference on Computer Vision and Pattern Recognition},
  2019.

\bibitem{bevformer}
Zhiqi Li, Wenhai Wang, Hongyang Li, Enze Xie, Chonghao Sima, Tong Lu, Yu Qiao,
  and Jifeng Dai.
\newblock Bevformer: Learning bird’s-eye-view representation from
  multi-camera images via spatiotemporal transformers.
\newblock {\em arXiv preprint arXiv:2203.17270}, 2022.

\bibitem{SSD}
Wei Liu, Dragomir Anguelov, Dumitru Erhan, Christian Szegedy, Scott Reed,
  Cheng-Yang Fu, and Alexander~C. Berg.
\newblock Ssd: Single shot multibox detector.
\newblock In {\em Proceedings of the European Conference on Computer Vision},
  2016.

\bibitem{CaDDN}
Cody Reading, Ali Harakeh, Julia Chae, and Steven~L. Waslander.
\newblock Categorical depth distributionnetwork for monocular 3d object
  detection.
\newblock {\em CVPR}, 2021.

\bibitem{imvoxelnet}
Danila Rukhovich, Anna Vorontsova, and Anton Konushin.
\newblock Imvoxelnet: Image to voxels projection for monocular and multi-view
  general-purpose 3d object detection.
\newblock In {\em Proceedings of the IEEE/CVF Winter Conference on Applications
  of Computer Vision}, pages 2397--2406, 2022.

\bibitem{Sun_2020_CVPR}
Pei Sun, Henrik Kretzschmar, Xerxes Dotiwalla, Aurelien Chouard, Vijaysai
  Patnaik, Paul Tsui, James Guo, Yin Zhou, Yuning Chai, Benjamin Caine, Vijay
  Vasudevan, Wei Han, Jiquan Ngiam, Hang Zhao, Aleksei Timofeev, Scott
  Ettinger, Maxim Krivokon, Amy Gao, Aditya Joshi, Yu Zhang, Jonathon Shlens,
  Zhifeng Chen, and Dragomir Anguelov.
\newblock Scalability in perception for autonomous driving: Waymo open dataset.
\newblock In {\em Proceedings of the IEEE/CVF Conference on Computer Vision and
  Pattern Recognition (CVPR)}, June 2020.

\bibitem{pgd}
Tai Wang, ZHU Xinge, Jiangmiao Pang, and Dahua Lin.
\newblock Probabilistic and geometric depth: Detecting objects in perspective.
\newblock In {\em Conference on Robot Learning}, pages 1475--1485. PMLR, 2022.

\bibitem{FCOS3D}
Tai Wang, Xinge Zhu, Jiangmiao Pang, and Dahua Lin.
\newblock {FCOS3D}: Fully convolutional one-stage monocular 3d object
  detection.
\newblock In {\em Proceedings of the IEEE/CVF International Conference on
  Computer Vision (ICCV) Workshops}, 2021.

\bibitem{DETR3D}
Yue Wang, Vitor Guizilini, Tianyuan Zhang, Yilun Wang, Hang Zhao, , and
  Justin~M. Solomon.
\newblock Detr3d: 3d object detection from multi-view images via 3d-to-2d
  queries.
\newblock In {\em The Conference on Robot Learning (CoRL)}, 2021.

\bibitem{SECOND}
Yan Yan, Yuxing Mao, and Bo Li.
\newblock Second: Sparsely embedded convolutional detection.
\newblock {\em Sensors}, 18(10), 2018.

\bibitem{CenterPoint}
Tianwei Yin, Xingyi Zhou, and Philipp Kr{\"a}henb{\"u}hl.
\newblock Center-based 3d object detection and tracking.
\newblock {\em CVPR}, 2021.

\end{thebibliography}
}

\end{document}